\documentclass[review]{elsarticle}
\usepackage{graphicx}
\usepackage{amssymb}
\usepackage{multirow}
\usepackage{mathptmx}
\usepackage{times}
\usepackage{epsfig}
\usepackage{graphicx}
\usepackage{amsmath}
\usepackage{amssymb}
\usepackage{multirow}
\usepackage[table,xcdraw]{xcolor}
\usepackage{caption}
\usepackage{todonotes}
\usepackage{xcolor}
\usepackage{url}
\usepackage{subfig}
\usepackage{blindtext}
\usepackage{lineno,hyperref}
\modulolinenumbers[5]

\journal{Journal of \LaTeX\ Templates}

\begin{document}

\begin{frontmatter}

\title{Breast Tumor Segmentation and Shape Classification in Mammograms using Generative Adversarial and Convolutional Neural Network}
\tnotetext[mytitlenote]{This work was supported by the Spanish project: DPI2016-77415-R and the Beatriu de Pin\'{o}s program: 2016-BP-00063.}


\author[mymainaddress]{Vivek Kumar Singh\corref{mycorrespondingauthor}}
\ead{vivekkumar.singh@urv.cat}

\author[mymainaddress]{Hatem A. Rashwan}
\ead{hatem.abdellatif@urv.cat}
\author[mymainaddress]{Santiago Romani}
\author[mysecondaryaddress]{Farhan Akram}
\author[mymainaddress]{Nidhi Pandey}
\author[mymainaddress]{Md. Mostafa Kamal Sarker}
\author[mymainaddress]{Adel Saleh}
\author[mythirdaddress]{Meritxell Arenas}
\author[mythirdaddress]{Miguel Arquez}
\author[mymainaddress]{Domenec Puig}
\author[mymainaddress]{Jordina Torrents-Barrena}

\cortext[mycorrespondingauthor]{Corresponding author}

\address[mymainaddress]{DEIM, Universitat Rovira i Virgili, Tarragona, Spain.}
\address[mysecondaryaddress]{Imaging Informatics Division, Bioinformatics Institute, A*STAR, Singapore.}
\address[mythirdaddress]{Hospital Universitari Sant Joan de Reus, Spain. }

\begin{abstract}
Mammogram inspection in search of breast tumors is
a tough assignment that radiologists must carry
out frequently. Therefore, image analysis methods are needed for the
detection and delineation of breast tumors, which portray crucial
morphological information that will support reliable diagnosis.
In this paper, we proposed a conditional Generative Adversarial
Network (cGAN) devised to segment a breast tumor within
a region of interest (ROI) in a mammogram. The generative
network learns to recognize the tumor area and to create
the binary mask that outlines it. In turn, the
adversarial network learns to distinguish between real (ground
truth) and synthetic segmentations, thus enforcing the generative
network to create binary masks as realistic as possible. The cGAN
works well even when the number of training samples are limited.
As a consequence, the proposed method outperforms several state-ofthe-art approaches. Our working hypothesis is corroborated by diverse
experiments performed on two datasets, the public INbreast and
a private in-house dataset. The proposed segmentation model
provides a high Dice coefficient and Intersection over Union (IoU)
of 94\% and 87\%, respectively. In addition, a shape descriptor
based on a Convolutional Neural Network (CNN) is proposed to
classify the generated masks into four tumor shapes: irregular,
lobular, oval and round. The proposed shape descriptor was
trained on Digital Database for Screening Mammography (DDSM)
yielding an overall accuracy of 80\%, which outperforms the
current state-of-the-art.
\end{abstract}


\end{frontmatter}


\section{Introduction}
\label{sec:Introduction}
Breast cancer is the most common diagnosed cause of death from cancer in women in the world \cite{siegel2017cancer}. Mammography is a world recognized tool that has been proven effective to reduce the mortality rate, since it allows early detection of breast diseases \cite{lauby2015breast}.

Breast masses are the most important findings among diverse types of breast abnormalities, such as micro-calcification and architectural distortion. All these findings may point out the presence of carcinomas \cite{rangayyan2010computer}. Moreover, morphological information of tumor shape (irregular, lobular, oval and round) and margin type (circumscribed, ill defined, spiculated and obscured) also play crucial roles in the diagnosis of tumor malignancy \cite{tang2009computer}.  

Computer aided diagnosis (CAD) systems are highly recommended to assist radiologists in detecting breast tumors and outlining their borders. However, breast tumor segmentation and classification are still challenges due to low signal-to-noise ratio and variability of tumors in shape, size, appearance, texture and location. 
Recently, many studies based on deep representation of breast images and combining features have been proposed to improve performance on breast mass classification \cite{jiao2018parasitic} .


In addition, based on mammographic images, it is very complicated for an expert radiologist to discern the molecular subtypes, i.e., Luminal-A, Luminal-B, HER-2 (Human Epidermal growth factor receptor 2) and Basal-like (triple negative), which are key for prescribing the best oncological treatment \cite{cho2016molecular}, \cite{liu2016there}, \cite{tamaki2011correlation}. However, recent studies point out some loose correlations between visual tumor features (e.g., texture and shape) and molecular subtypes. Recently, a Convolutional  Neural Network (CNN) was used to classify molecular subtypes using texture patches extracted from mammography \cite{singh2017classification}, which yielded an overall accuracy of 67\%.  However, depending only on texture feature is not sufficient to classify the breast cancer molecular subtypes from mammograms \cite{tamaki2011correlation}. Thus, some studies attempt to use morphological information of tumor shape in classifying breast cancer molecular subtypes.

\begin{figure*}[htp]
\centering
\includegraphics[width=1.0\textwidth, height=0.17\textheight]{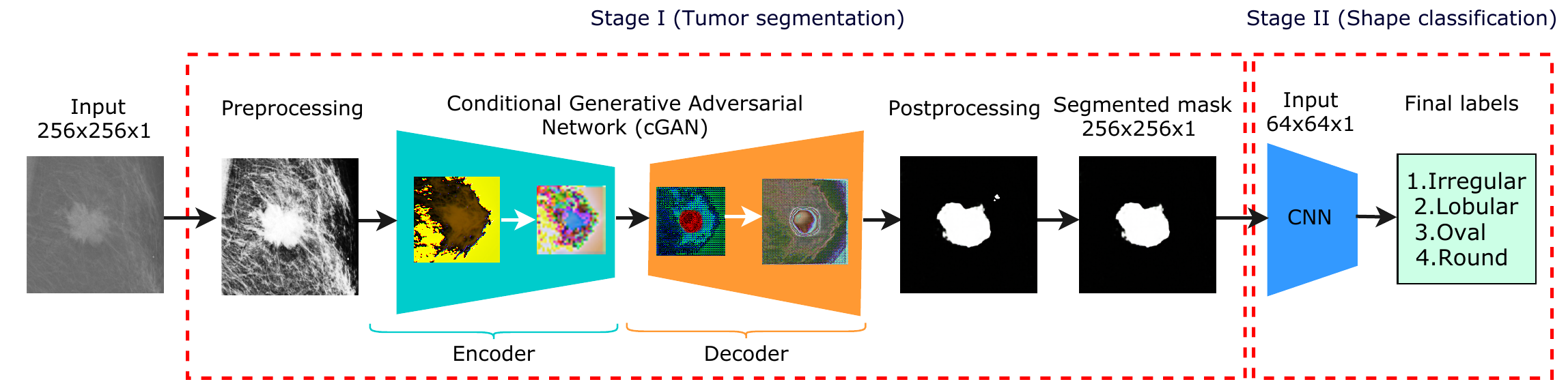}
\caption{General framework of breast tumor segmentation and shape classification.}
\label{fig:globalmodel}
\end{figure*}



Consequently, in this paper, a method of two stages of breast tumor segmentation and shape classification is proposed as shown in Figure~\ref{fig:globalmodel}. In the first stage, our method segments the breast tumor as a binary mask. In the second stage, the binary mask is classified to a shape type (irregular, lobular, oval and round). Unlike traditional object classifiers \cite{kisilev2015semantic}, \cite{ kim2018icadx} that use texture, intensity or edge information, our method is forced to learn only morphological features from the binary masks. To be more specific, we present a thorough improvement of our previous work \cite{singh2018conditional}. The major contributions of this paper are as follows:

\vspace{0.5mm}
\begin{enumerate}
\item We believe this is the first adaptation of cGAN in the area of breast tumor segmentation in mammograms. The adversarial network yields more reliable learning than other state-of-the-art algorithms since training data is scarce (\emph{i.e.,} mammograms with labeled breast tumor boundaries), while it does not increase the computational complexity at prediction time.\vspace{2.mm}

\item The implementation of a multi-class CNN architecture to predict the four breast tumor shapes (\emph{i.e.,} irregular, lobular, oval and round) using the binary mask segmented in the previous stage (cGAN output).\vspace{2.mm}

\item An in-depth evaluation of our system's performance using two public (1,274 images) and one private (300 images) databases. The obtained results outperform current state-of-the-art in both tumor segmentation and shape classification.

\item A study of the correlation between the tumor shape and molecular subtypes of breast cancer is also provided.
\end{enumerate}
\vspace{1.5mm}

This paper is organized as follows. Section II provides the related work of both tumor segmentation and shape classification. The proposed architectures for tumor segmentation (using cGAN) and shape classification (using CNN) are described in Section III. In Section IV, extensive experiments are performed on the two stages of the proposed method and the obtained results are compared with the state-of-the-art results. In addition, the limitations of the proposed models are explained in Section IV.  Finally, Section V concludes our work and suggests some future lines of research.


\section{Related Work}
\label{sec:relwork}

\subsection{Tumor Segmentation Background}
\label{subsec:tumorSeg}
Convolutional Neural Networks (CNNs) can automatically learn features from the given images to represent objects at different scales and orientations. By increasing the number of layers (depth of CNN model) more detailed features can be obtained, which play crucial part in solving different computer vision problems, such as object detection, classification and segmentation. Thus, numerous methods has been proposed to solve the image segmentation problem based on deep learning approaches~\cite{schmidhuber2015deep}. One of the well-known architectures for semantic segmentation is the Fully Convolutional Network (FCN) \cite{long2015fully}, which is based on encoding (convolutional) and decoding (deconvolutional) layers. This approach gets rid of the fully connected layers of CNNs to convert the image classification networks into image filtering networks. An improvement of this scheme was proposed by the U-Net architecture \cite{ronneberger2015u}, where skip connections between encoding and decoding layers are added to retain significant information from the input features. Later on, a new variation of FCN was proposed \cite{badrinarayanan2017segnet} named SegNet, which consists of hierarchy of decoders, each one corresponding to each encoder. The decoder network uses the max-pooling indices received from the corresponding encoder to perform non-linear upsampling of their input feature maps.

Since semantic segmentation has achieved great progress with deep learning, there is recent popularity in applying such models to medical imaging \cite{litjens2017survey}. For instance, to segment skin lesions on dermoscopic images, the SLSDeep model \cite{sarker2018slsdeep} was proposed to upscale the feature maps from the encoding layers at multi-scale to preserve small details (\emph{e.g.}, lesion borders). In~\cite{fu2018joint}, a multi-scale deep model with multi-level loss was proposed for segmenting optic disk and cup in Fundus images. Also, \cite{singh2018retinal} proposed a GAN to segment the optic disc from fundus image. Many segmentation approaches can be trained from scratch~\cite{tajbakhsh2016convolutional} but also can reuse the weights obtained for the starting CNN layers of other architectures such as ResNet~\cite{he2016deep} and VGG \cite{simonyan2014very} trained on ImageNet data \cite{deng2009imagenet}.

Regarding breast tumor segmentation, many works have been proposed. A tumor classification and segmentation method was proposed \cite{rouhi2015benign} using an automated region growing algorithm whose threshold was obtained by a trained Artificial Neural Network (ANN) and Cellular Neural Network (CeNN). In turn, to reduce the computational complexity and increase the robustness, a quantized and non-linear CeNN for breast tumor segmentation was proposed in~\cite{liu2018efficient}. After segmenting the breast tumor region, a Multilayer Perceptron Classifier was used for tumor classification as benign or malignant.

Furthermore, Dhungel et al. \cite{dhungel2015deep} segmented breast tumors using Structured Support Vector Machines (SSVM) and Conditional Random Fields (CRF). Both graphical models minimize a loss function build on pixel probabilities provided by a CNN and Deep Belief Network, a Gaussian Mixture Model (GMM) and shape prior. The SSVM is based on graph cuts and the CRF relies on tree re-weighted belief propagation with truncated fitting training ~\cite{dhungel2015tree}. Cardoso et al. \cite{cardoso2015,cardoso2017mass} tackled the same problem by employing a closed contour fitting in the mammogram and minimizing a cost function depending on the radial derivative of the tumor contour. A measure of regularity of the gray pixel values inside and outside the tumor was also included in \cite{cardoso2017mass}.

In turn, Zhu et al. \cite{zhu2018adversarial} proposed an FCN concatenated to a CRF layer to impose the compactness of the segmentation output taking into account pixel position. This approach was trained end-to-end, since the CRF and FCN can exchange data in the forward-backward propagation. An adversarial term was introduced to prevent the samples with the worst perturbation in the loss function, which reduced the overfitting and provided a robust learning with few training samples. In addition, Al-antari et al. \cite{al2018fully} proposed a CAD system consisting of three deep learning stages for detecting, segmenting and classifying the tumors in mammographic images. To locate tumors in a full mammogram, the YOLO network proposed in~\cite{redmon2016you} was used. A Full resolution Convolutional Network (FrCN) was then used for segmenting the located tumor region. Finally, a CNN network was used for classifying segmented ROI as either benign or malignant.

We believe that \cite{yang2017automatic} is the first work that exploits GAN \cite{goodfellow2014} for medical image segmentation. In particular, they performed three-dimensional (3D) liver segmentations using abdominal computerized tomography (CT) scans. In~\cite{singh2018conditional}, we adapted a cGAN image-to-image translation algorithm \cite{isola2017image} to address the tumor segmentation in two-dimensional (2D) mammograms. Our system provided state-of-the-art performance on both public and private databases.

\subsection{Shape Classification Background}
\label{subsec:shapeClass}
In the literature, many approaches based on deep learning architectures have been designed recently for 2D and 3D shape classification \cite{kurnianggoro2018survey}. For example, topological data analysis (TDA) using deep learning was proposed in \cite{hofer2017deep} to extract relevant 2D/3D topological and geometrical information. In turn, a CNN model was formulated, which used spectral graph wavelets in conjunction with the Bag of Features paradigm to target the shape classification problem \cite{masoumi2017}. In addition, the authors in~\cite{fang20153d} proposed a CNN based shape descriptor for retrieving the 3D shapes. A deep neural network named PointNet was proposed \cite{qi2017pointnet}, which directly consumes point cloud for object classification, localized and global semantic segmentation. Moreover, a deep learning framework for efficient 3D shape classification \cite{luciano2018deep} used geodesic moments by inheriting various properties from the geodesic distance, such as the intrinsic geometric structure of 3D shapes and the invariance to isometric deformations.

To date, numerous shape classification methods are applied for medical image analysis \cite{litjens2017survey}. Fourier shape descriptors with a CNN were used \cite{xie2018fusing} to characterize the lung nodules heterogeneity in CT scans. A CNN architecture coupled with neighboring ensemble predictor invariant to the neighborhood was proposed \cite{sirinukunwattana2016locality} for nucleus detection and classification in histological images.


An automated method for textual description of anatomical breast tumor lesions was proposed by Kisilev et al. \cite{kisilev2015semantic}, which performs joint semantic estimation from image measurements to classify the tumor shape. In addition, Kisilev et al. \cite{kisilev2016medical} also presented a multi-task fast region-based CNN \cite{ren2015faster} to classify three tumor shapes: irregular, oval and round. Furthermore, the work in \cite{kim2018icadx} utilized a GAN to diagnose and classify tumors in mammograms into four shapes: irregular, lobular, oval and round. Previously, Singh et al. \cite{singh2018conditional} proposed a multi-class CNN to categorize the tumor shapes into four classes as in \cite{kim2018icadx} from the public dataset DDSM\footnote{https://wiki.cancerimagingarchive.net/display/Public/CBIS-DDSM}.

\section{Proposed Methodology}
\label{sec:method}
The proposed CAD system shown in Fig.~\ref{fig:globalmodel} is divided into two stages: breast tumor segmentation and shape classification. In the first stage, mammograms are pre-processed for noise removal (Gaussian filter with $\sigma$ = 0.5) and then contrast is enhanced using histogram equalization (pixel values are rescaled between [0..1]). Afterwards, the cGAN input is prepared by rescaling the image crops to 256$\times$256 pixels containing different framing of the breast tumor region (ROI): full mammogram, loose and tight frames (see Fig.~\ref{fig:croppings}). The prepared data is then fed to the cGAN to obtain a binary mask of the breast tumor, which is post-processed using morphological operations (3$\times$3 closing, 2$\times$2 erosion, and 3$\times$3 dilation) to remove small speckles. In the second stage, the output binary mask is downsampled into 64$\times$64 pixels, which is then fed to a multi-class CNN shape descriptor to categorize it into four classes: irregular, lobular, oval and round.

\subsection{Tumor Segmentation Model (cGAN)}
\label{subsec:tumorSegcGAN}
\begin{figure*}[htp]
\centering
\includegraphics[width=1.0\textwidth, height=0.45\textheight]{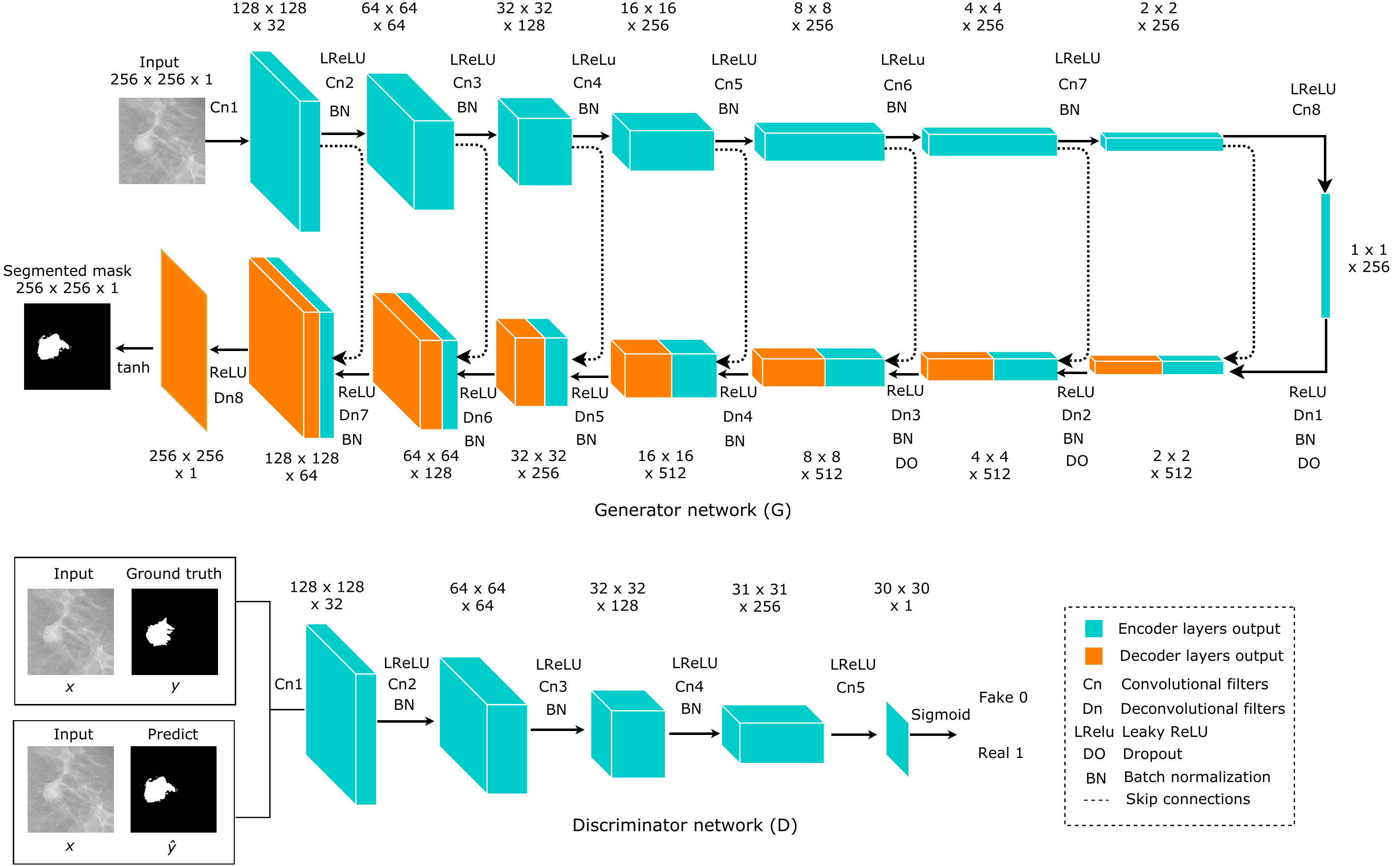}
\caption{Proposed cGAN architecture: generator \emph{G} (top), and discriminator \emph{D} (down).}
\label{fig:cGAN architecture}
\end{figure*}

Our previous work \cite{singh2018conditional} demonstrated the feasibility of applying the cGAN image-to-image translation approach \cite{isola2017image} to breast tumor segmentation, since it can be adapted to our problem in the following senses:
\vspace{1.5mm}

\begin{enumerate}
  \item The Generator $G$ network of the cGAN is an FCN composed of encoding and decoding layers, which learn the intrinsic features (gray-level, texture, gradients, edges, shape, etc.) of healthy and unhealthy (tumor) breast tissue, and generate a binary mask according to these features.\vspace{1.5mm}
  \item The Discriminative $D$ network of the cGAN assesses if a given binary mask is likely to be a realistic segmentation or not. Therefore, including the adversarial score in the computation of the generator loss strengthens its capability to provide a correct segmentation.
\end{enumerate}
\vspace{1.5mm}

The combination of $G$ and $D$ networks allows robust learning with few training samples. Since the ROI image is a conditioning input for both $G$ and $D$, the segmentation result is better fitted to the tumor appearance. Otherwise, regular (unconditional) GAN \cite{goodfellow2014} will infer the segmentation just from random noise, which will require more training iterations compared to the cGAN to obtain an acceptable segmentation result.

Fig.~\ref{fig:cGAN architecture} represents the suggested architectures for $G$ and $D$. The former consists of several encoding and decoding layers (see Fig.~\ref{fig:cGAN architecture}-top). Encoding layers are composed of a set of convolutional filters followed by batch normalization and the leaky ReLU (slope $0.2$) activation function. Similarly, decoding layers are composed of a set of deconvolutional filters followed by batch normalization, dropout and ReLU.

Convolutional and deconvolutional filters are defined with a kernel of 4$\times$4 and stride of 2$\times$2, which respectively downsample and upsample the activation maps by a factor of 2. Batch normalization is not applied after the first and the last convolutional filters ($Cn_1$ and $Cn_8$). After $Cn_8$, the ReLU activation function is applied instead of  leaky ReLU. Dropout is applied only at the first three decoding layers ($Dn_1$, $Dn_2$ and $Dn_3$). There is no skip connection in the last decoding layer ($Dn_8$), after which the $tanh$ activation function is applied to generate a binary mask of the breast tumor.


The architecture of $D$ shown in Fig.~\ref{fig:cGAN architecture}-down consists of five encoding layers with convolutional filters with a kernel of 4$\times$4, stride 2$\times$2 at the first three layers and stride 1$\times$1 at 4$^{th}$ and 5$^{th}$ layers. Batch normalization is applied after $Cn_2$, $Cn_3$ and $Cn_4$ and a leaky ReLU (slope $0.2$) is applied after each layer except for the last one. The sigmoid activation function is used after the last convolutional filter ($Cn_5$). The network input is the concatenation of the ROI and the binary mask to be evaluated (ground truth or predicted). The output segmentation is an array of 30$\times$30 values, each one from $0.0$ (completely fake) to $1.0$ (perfectly plausible or real). Each output value is the degree of proper segmentation likelihood of a crop of the binary mask and the input image, which corresponds to a 70$\times$70 receptive field for each value.

\begin{figure}[htp]
\centering
\includegraphics[width=0.7\textwidth, height=0.3\textheight]{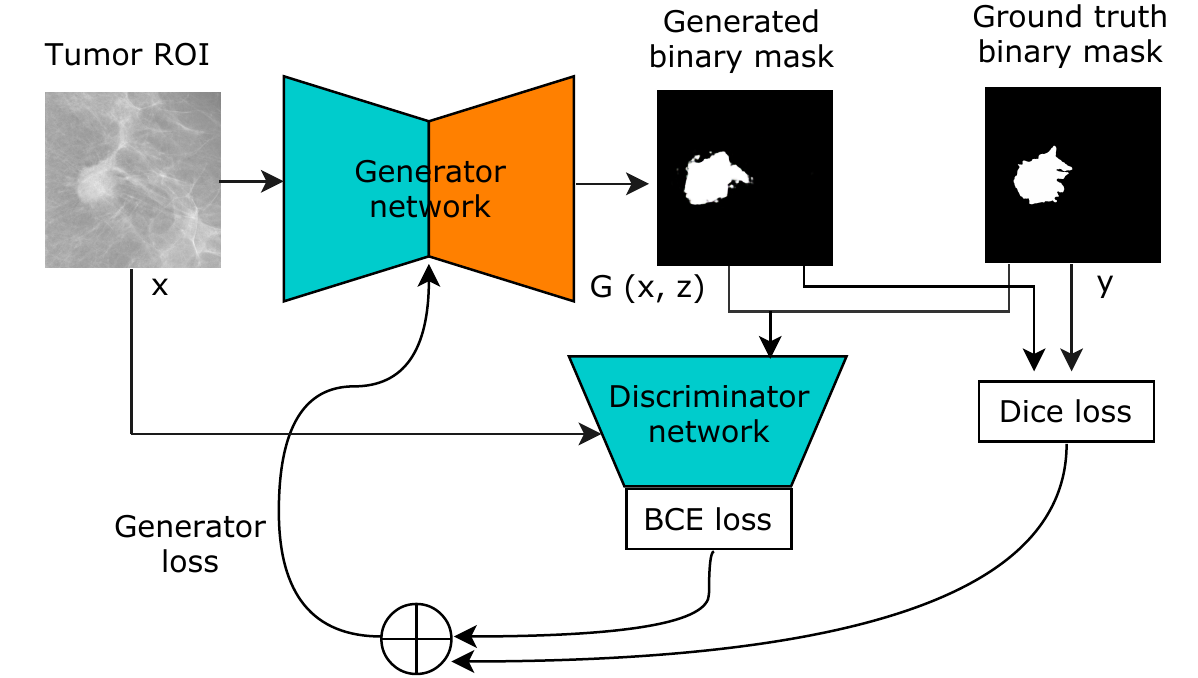}
\caption{Proposed cGAN framework based on dice and BCE losses.}
\label{fig:generator loss}
\end{figure}

Let $x$ be a tumor ROI, $y$ the ground truth mask, $z$ a random variable, $\lambda$ an empirical weighting factor, $G(x, z)$ and $D(x, G(x,z))$ the outputs of $G$ and $D$, respectively. Then, the loss function of $G$ is defined as:

\begin{align}
\label{Lgenupdate}
\ell_{Gen}(G, D) &=  \mathbb{E}_{x,y,z}(-\log(D(x, G(x,z))))+ \lambda \mathbb{E}_{x,y,z}(\ell_{Dice}(y, G(x,z))),
\end{align}
where $z$ is introduced as dropout in the decoding layers $Dn_1$, $Dn_2$ and $Dn_3$ at both training and testing phases, which provides stochasticity to generalize the learning processes and avoid overfitting.

The optimization process of $G$ will try to minimize both expected values, \emph{i.e.}, the $D$ values should approach to $1.0$ (correct tumor segmentations), and the dice loss $\ell_{Dice}$ should approach to $0.0$ (generated masks are equal to ground truth). Both terms of generator loss enforce the proper optimization of $G$: the dice loss term fosters a rough prediction of the mask shape (central tumor area) while the adversarial term fosters an accurate prediction of the mask outline (tumor borders). Neglecting one of the two terms may lead to either very poor segmentation results or slow learning speed.

In addition, $\ell_{Dice}(y, G(x,z))$ is the dice loss of the predicted mask with respect to ground truth, which is defined as:

\begin{align}
\label{dice}
\ell_{Dice}(y, z) = 1 - \frac{2|y \circ G(x,z)|}{|y|+|G(x,z)|},
\end{align}
where $\circ$ is the pixel wise multiplication of the two images and $|.|$ is the total sum of pixel values of a given image. If inputs are binary images, then each pixel can be considered as a boolean value (white is $1$ / black is $0$ ). The formulation in (\ref{dice}) is equivalent to the dice coefficient \emph{i.e.,} $2\times\frac{TP}{TP+FN+TP+FP}$, but it must be subtracted from $1.0$ because the loss function will be minimized. Let $A$ be the ground truth of the ROI and $B$ the segmented region. Then the true positive degree (TP) is defined as $TP=A\cap B$, which is the area of the segmented region common in both $A$ and $B$. The false positive degree (FP) is defined as $\overline{A} \cap B$, which is the segmented area not belonging to $A$. Similarly, the false negative degree (FN) is defined as $A \cap \overline{B}$, which is the true area missed by the proposed segmentation method.

In our previous work \cite{singh2018conditional}, the generator network loss was formulated by combining the logistic Binary Cross Entropy (BCE) loss and the $L1$-norm. In this work, we replace the $L1$-norm loss with the dice loss as shown in Fig.~\ref{fig:generator loss}. $L1$-norm loss minimizes the sum of absolute differences between the ground truth label $y$ and estimated binary mask $G(x,z)$ obtained from the generator network, which takes all pixels into account. In turn, dice loss is highly dependent on TP predictions, which is the most influential term in foreground segmentation. Fig.~\ref{fig:lossfunction} shows that the dice loss achieves lower values (more optimal) than the $L1$-norm loss.

\begin{figure}[htp]
\centering
\includegraphics[width=0.8\textwidth, height=0.35\textheight]{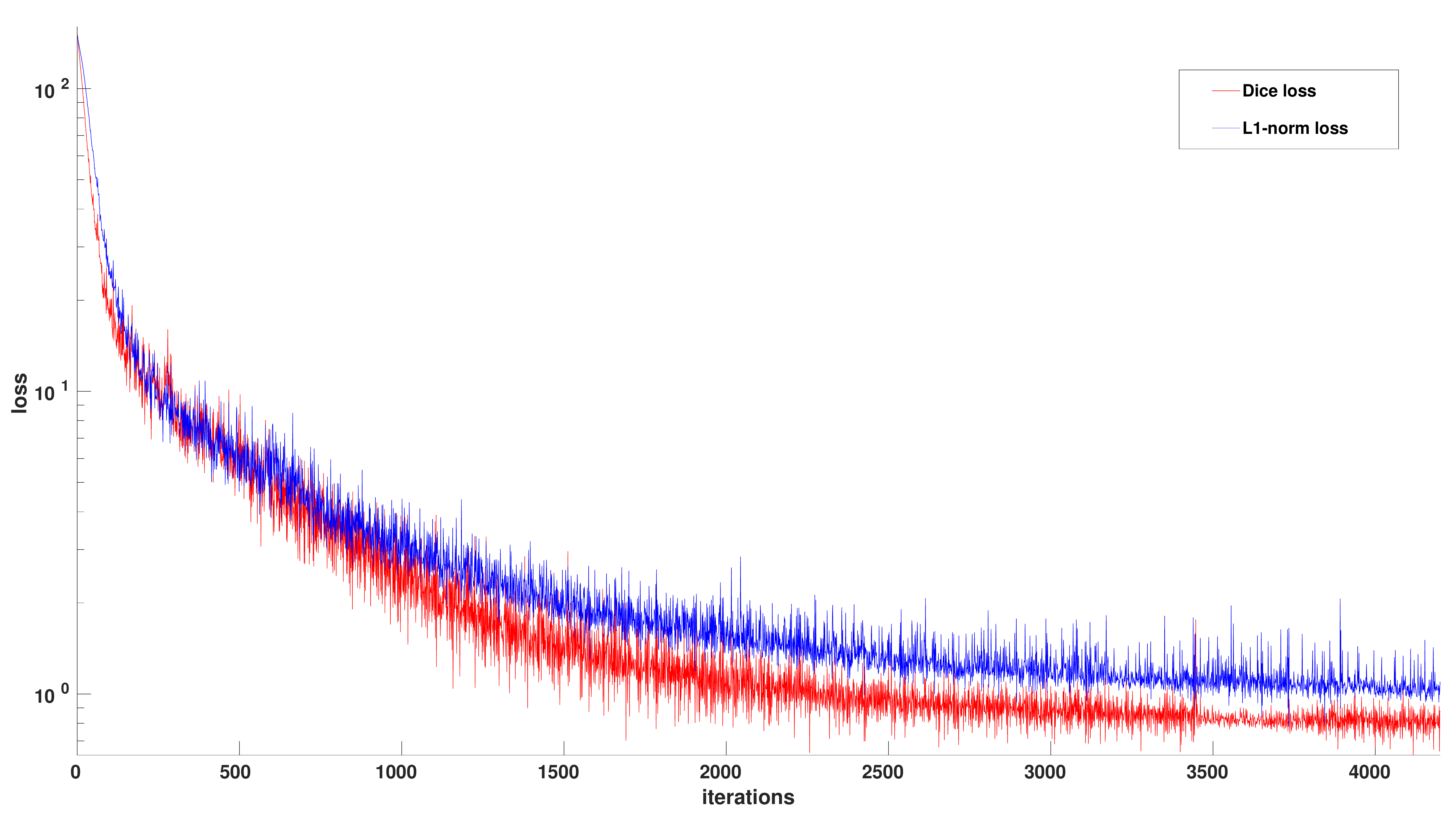}
\caption{Dice and $L$1-norm loss comparison over iterations.}
\label{fig:lossfunction}
\end{figure}

Moreover, the loss function of $D$ is defined in (\ref{Ldis}):

\begin{align}
\label{Ldis}
\ell_{Dis}(G, D) &= \mathbb{E}_{x,y,z}(-\log(D(x, y))) + \mathbb{E}_{x,y,z}(-\log(1-D(x, G(x,z))))
\end{align}

The optimizer will fit $D$ to maximize the loss values for ground truth masks (by minimizing $-\log(D(x, y))$) and minimize the loss values for generated masks (by minimizing $-\log(1-D(x, G(x, z))$). These two terms compute BCE loss using both masks, assuming that the expected class for ground truth and generated masks is $1$ and $0$, respectively.

The optimization of $G$ and $D$ is done concurrently, \emph{i.e.}, one optimization step for both networks at each iteration, where $G$ learns how to compute a valid tumor segmentations and $D$ learns how to differentiate between synthetic and real segmentations.

In this work, we experimented on different hyper-parameters to improve the segmentation accuracy of our previous contribution in \cite{singh2018conditional}. Besides introducing the dice loss, we have reduced the number of filters of each network from 64 to 32. We also explored different learning rates and loss optimizers (SGD, AdaGrad, Adadelta, RMSProp and Adam), finding Adam with $\beta_1$ = 0.5, $\beta_2$ = 0.999 and initial learning rate = 0.0002 with batch size 8 the best combination. In (\ref{Lgenupdate}), the dice loss weighting factor $\lambda=150$ was found to be the best choice. Finally, the best results were achieved by training both $G$ and $D$ from scratch for 150 epochs.

\subsection{Shape Classification Model (CNN)}
\label{subsec:shapeCNN}
We propose a multi-class CNN architecture for breast tumor shape classification (\emph{i.e.}, irregular, lobular, oval and round) using the binary masks obtained from the cGAN. In the literature, most methods attempted to directly categorize the shape using breast tumor intensity, texture, boundary, etc. (\cite{kisilev2015semantic, kisilev2016medical, ren2015faster, kim2018icadx}), which increase computational complexity. We simplify the problem by extracting morphological features from binary masks.

\begin{figure*}[htp]
\centering
\includegraphics[width=1.0\textwidth, height=0.20\textheight]{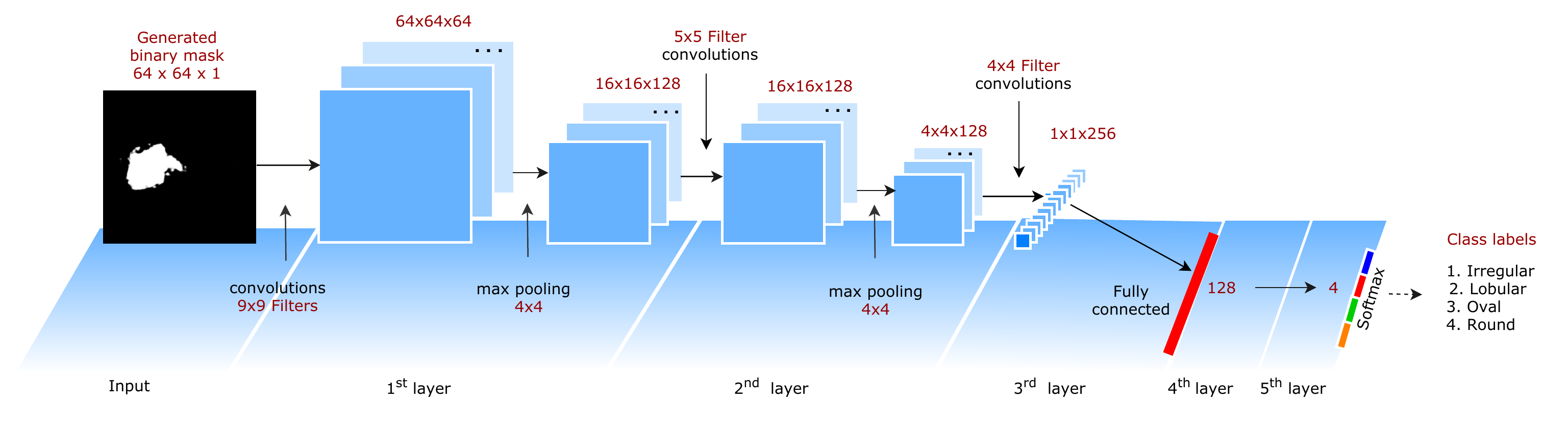}
\caption{CNN architecture for tumor shape classification.}
\label{fig:CNN architecture}
\end{figure*}

As shown in Fig. \ref{fig:CNN architecture}, our model consists of three convolutional layers with kernel sizes 9$\times$9, 5$\times$5 and 4$\times$4, respectively, and two fully connected (FC) layers. The first two convolutional layers are followed by 4$\times$4 max-pooling with stride 4$\times$4. The output of the last convolutional layer is flattened and then fed into the first FC layer with 128 neurons. These four layers use ReLU as activation function. A dropout of 0.5 is used to reduce overfitting in the first FC layer. Finally, the last FC layer with 4 neurons applies the softmax function to generate the final membership degree of the input binary mask to each class. A weighted categorical cross-entropy loss is used to avoid the problem of unbalanced dataset. The class weight is one minus the ratio of samples per class to the total number of samples.

The RMSProp is employed for optimizing the model with learning rate = 0.001, momentum = 0.9 and batch size = 16. The network is trained from scratch and the weights of five layers are randomly initialized. During training, we experimentally found the best architecture, number of layers, filters per layer, and number of neurons in FC layers.

\section{Experiments and Discussion}
\label{sec:expDis}
We have evaluated the performance of proposed models on two public mammography datasets and one private dataset:

\subsubsection*{\textbf{INbreast dataset\footnote{\url{http://medicalresearch.inescporto.pt/breastcancer/index.php/Get_INbreast_Database/}}}}
It is a publicly available database containing a total of $115$ cases ($410$ mammograms), which include: masses, calcifications, asymmetries and distortions. For testing our segmentation model, we used $106$ breast tumor images along with their respective ground truth binary masks.

\subsubsection*{\textbf{DDSM dataset}}
It is a publicly available digital database for screening mammography containing $2,620$ mammography studies. 
In this work, $1,168$ cases of breast tumors with their corresponding ground truths are used for shape classification, where $504$, $473$, $115$ and $76$ tumors are labeled as irregular, lobular, oval and round, respectively. We have used $75\%$ of the images for training and rest for testing the tumor shape classification model.


\subsubsection*{\textbf{Hospital Sant Joan de Reus dataset}}
It is our private dataset that contains $300$ malignant tumors (123 Luminal-A, 107 Luminal-B, 33 Her-2
and 37 Basal-like) with their respective ground truth binary masks obtained by radiologists. The proposed cGAN segmentation model is trained and tested using $220$ and $80$ images, respectively. The duty of confidentiality and security measures were fully complied, in accordance with the current legislation on the Protection of Personal Data (article 7.1 of the Organic Law 15/1999, 13th of December).

The proposed method was implemented using python with Pytorch\footnote{\url{https://pytorch.org/}}  running on a 64-bit Ubuntu operating system using a 3.4 GHz Intel Core-i7 with 16 GB of RAM and Nvidia GTX 1070 GPU with 8 GB of video RAM.

\subsection{Tumor Segmentation Experiments}
The proposed breast tumor segmentation method is compared with the state-of-the-art methods and evaluated both quantitatively and qualitatively. For the quantitative analysis, segmentation accuracy is computed using Dice coefficient (F1 score) and Jaccard index (IoU). In turn, for the qualitative analysis, segmentation results with the their respective ground truth binary masks are compared visually.

\begin{figure}[htp]
\centering
\includegraphics[width=0.6\textwidth, height=0.2\textheight]{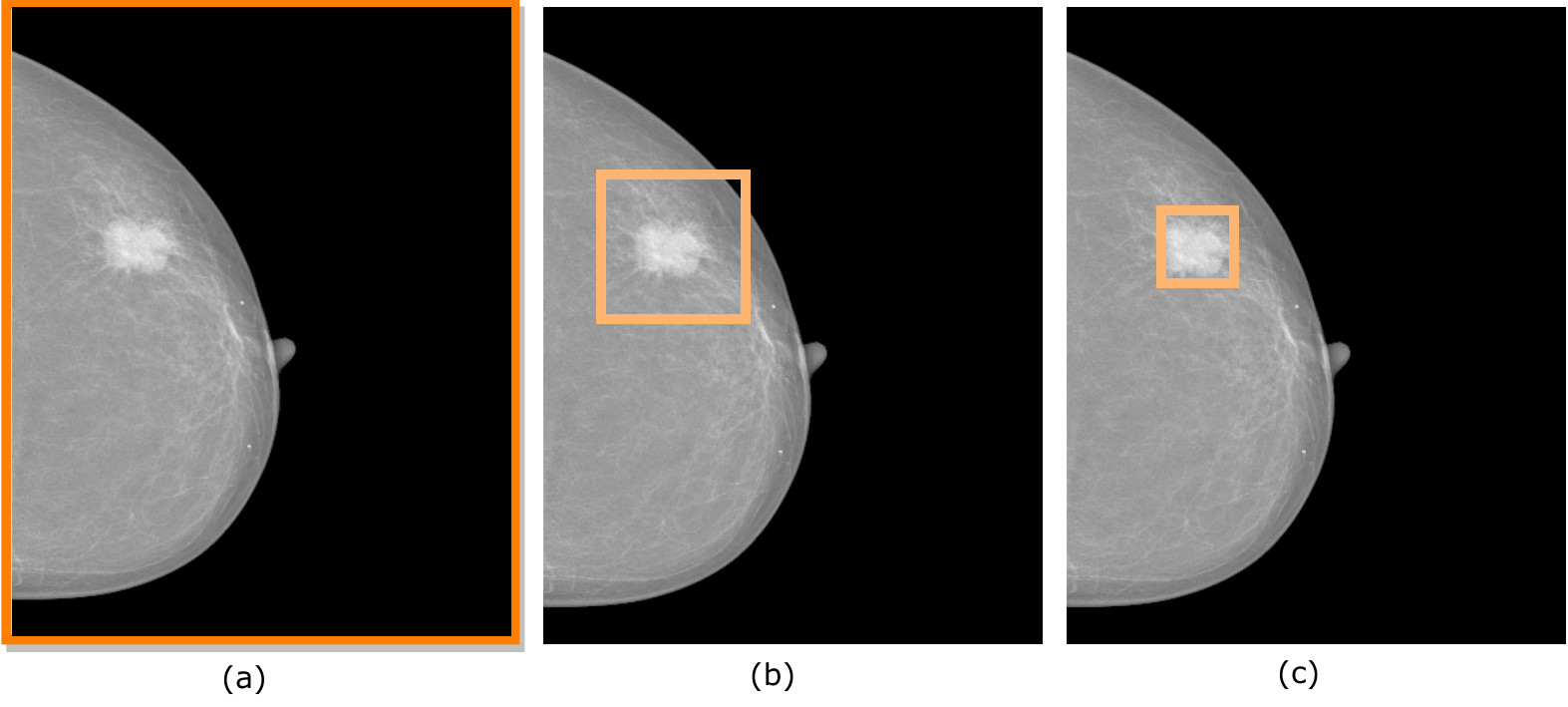}
\caption{Three cropping strategies: (a) full mammogram, (b) loose frame, (c) tight frame.}
\label{fig:croppings}
\end{figure}

\begin{figure}[htp]
\centering
\includegraphics[width=0.6\textwidth, height=0.4\textheight]{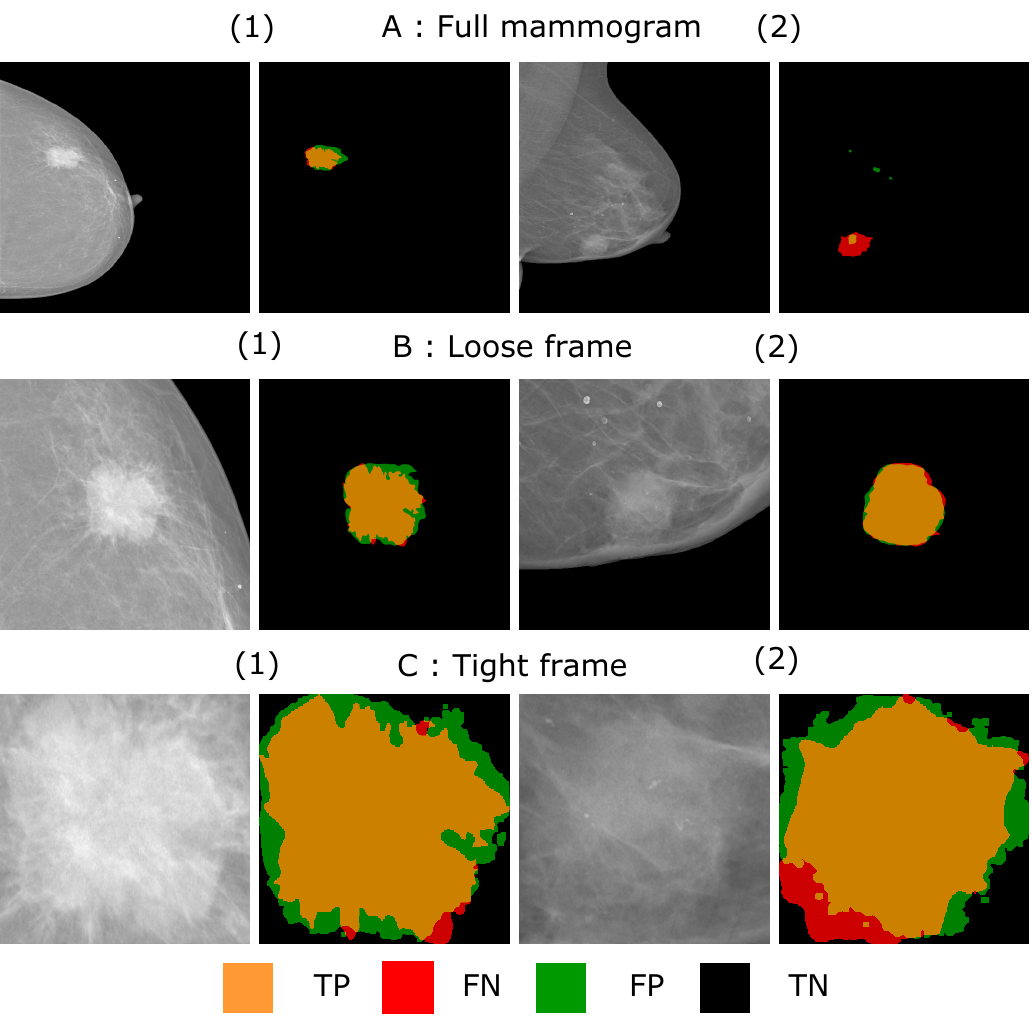}
\caption{Segmentation results of two testing samples extracted from the INbreast dataset with the three cropping strategies.}
\label{fig:segment2x3examples}
\end{figure}

These experiments have been carried using three different framing of the tumor ROI: full mammogram, loose and tight frames (see Fig.~\ref{fig:croppings}). The ideal CAD system should be able to automatically segment the breast tumor from a full mammogram. However, this is a very difficult task due to high similarity between gray level pixel distributions of healthy and tumorous tissue. Therefore, removing most of non-ROI portions of the image logically helps the model on learning the visual features that differentiate breast tumor from non-tumor areas. The loose frame provides a balanced proportion between the number of pixels of the two classes. The tight frame is intended to evaluate the behavior of the segmentation model when the majority of ROI contains tumor pixels. Experimentally, for detecting the tight frame, we used the deep model Single Shot Detector (SSD), recently proposed in~\cite{liu2016ssd}. In turn, the loose frame is selected by doubling the size of the tight frame in each coordinate (see Fig.~\ref{fig:segment2x3examples}).


The three cropping strategies are evaluated on our cGAN and ten baseline segmentation models, referred as FCN, FCN-ResNet101, UNet, UNet-VGG16, SegNet, SegNet-VGG16, CRFCNN, SLSDeep, cGAN-ResNet101 and  cGAN-ResNet101 (Dice Loss). FCN, UNet, SegNet, CNNCRF and proposed cGAN are trained from scratch. FCN-ResNet101, UNet-VGG16, SegNet-VGG16 and cGAN-ResNet101 (with and without Dice loss) are modifications of the original models, where the filters of the starting encoding layers are replaced by the starting convolutional layers of the well-known VGG (16 layers) and ResNet (101 layers) models, which were pre-trained on the ImageNet database. Thus, we loaded the pre-trained weights and fine tuned the network. When using cGAN-ResNet101~\cite{isola2017image}, we replaced the $L1$-norm loss with the Dice loss in the generator loss function to see how the base line model will behave under such change. We called this model cGAN-ResNet101 (Dice loss) to compare the segmentation results with our proposal.

\begin{table}[htp]
\centering
\caption{ Dice and IoU metrics obtained with the proposed model and ten alternatives evaluated on the testing sets of our private and INbreast datasets, for the three cropping strategies. Best results are marked in bold. Dashes (-) indicate that results are not reported in referred papers.}
\label{tabDiceIoU}
\scalebox{0.9}{
\begin{tabular}{c|c|ccc|ccc}
\hline 
\multicolumn{1}{c|}{}     &
& \multicolumn{3}{c|}{ Dice (\%)}                                         
& \multicolumn{3}{c}{IoU (\%)}          \\\cline{3-8}
{Dataset}     &  Methods             
& { Full} & { Loose} &{ Tight}  & { Full} & {Loose} & { Tight} \\ \hline 
\multicolumn{1}{c|}{}                           & FCN                                    & $59.06$                                               & $74.94$                                                & $80.20$                                                & $39.92$                                              & $62.21$                                              & $78.89$         \\
\multicolumn{1}{c|}{}                           & FCN-ResNet101                             & $59.21$                                               & $77.42$                                                & $82.78$                                                & $40.26$                                              & $68.16$                                              & $77.32$         \\
\multicolumn{1}{c|}{}                           & UNet                                      & $63.69$                                               & $78.03$                                                & $83.15$                                                & $46.73$                                              & $68.36$                                              & $78.81$         \\
\multicolumn{1}{c|}{}                           & UNet-VGG16                                  & $59.27$                                               & $78.57$                                                & $83.71$                                                & $42.13$                                              & $69.71$                                              & $79.42$         \\
\multicolumn{1}{c|}{}                           & SegNet                                     & $59.87$                                               & $80.26$                                                & $82.33$                                                & $42.79$                                              & $70.07$                                              & $76.17$         \\
\multicolumn{1}{c|}{}                           & SegNet-VGG16                                 & $61.59$                                               & $81.09$                                                & $81.41$                                                & $41.61$                                              & $68.19$                                              & $77.82$         \\
\multicolumn{1}{c|}{}                           & CRFCNN                                     & $53.21$                                               & $71.33$                                                & $63.52$                                                & $41.38$                                              & $65.24$                                              & $54.28$         \\
\multicolumn{1}{c|}{}                           & SLSDeep                                    & $59.64$                                               & $71.10$                                                & $84.28$                                                & $43.89$                                              & $60.16$                                              & \textbf{79.93}         \\
\multicolumn{1}{c|}{}                           & cGAN-ResNet101                           & $58.37$                                               & $80.11$                                                & $86.22$                                                & $42.12$                                              & $71.91$                                              & $76.62$         \\

\multicolumn{1}{c|}{}                           & cGAN-ResNet101 (Dice Loss)                           & $61.49$                                               & $86.57$                                                & $86.37$                                                & $45.90$                                              & $76.32$                                              & $77.26$         \\

\multicolumn{1}{c|}{\multirow{-11}{*}{Private}} & \textbf{Proposed cGAN}               & \textbf{66.38}                                               & \textbf{89.99}                                                & \textbf{88.12}                                                & \textbf{49.68}                                              & \textbf{81.81}                                              & $79.87$         \\ \hline
\multicolumn{1}{c|}{}                           & FCN                                    & $54.36$                                               & $66.12$                                                & $81.74$                                                & $36.88$                                              & $49.38$                                              & $77.33$         \\
\multicolumn{1}{c|}{}                           & FCN-ResNet101                             & $51.76$                                               & $83.80$                                                & $82.38$                                                & $38.49$                                              & $74.12$                                              & $78.09$         \\
\multicolumn{1}{c|}{}                           & UNet                                      & $55.58$                                               & $77.92$                                                & $80.76$                                                & $38.46$                                              & $70.83$                                              & $77.97$         \\
\multicolumn{1}{c|}{}                           & UNet-VGG16                                  & $56.79$                                               & $78.02$                                                & $80.89$                                                & $39.65$                                              & $68.32$                                              & $78.13$         \\
\multicolumn{1}{c|}{}                           & SegNet                                     & $53.33$                                               & $79.06$                                                & $81.11$                                                & $36.36$                                              & $65.37$                                              & $77.02$         \\
\multicolumn{1}{c|}{}                           & SegNet-VGG16                                 & $56.27$                                               & $80.17$                                                & $81.75$                                                & $39.46$                                              & $69.79$                                              & $78.68$         \\
\multicolumn{1}{c|}{}                           & CRFCNN                                     & $52.96$                                               & $73.25$                                                & $65.41$                                                & $40.41$                                              & $67.14$                                              & $57.69$         \\
\multicolumn{1}{c|}{}                           & SLSDeep                                    & $60.35$                                               & $75.90$                                                & $85.53$                                                & $44.63$                                              & $65.16$                                              & $80.26$         \\
\multicolumn{1}{c|}{}                           & cGAN-ResNet101                           & $54.69$                                               & $87.19$                                                & $89.17$                                                & $37.94$                                              & $77.51$                                              & $82.26$         \\

\multicolumn{1}{c|}{}                           & cGAN-ResNet101 (Dice Loss)                           & $59.72$                                               & $88.89$                                                & $90.42$                                                & $44.89$                                              & $82.58$                                              & $82.95$         \\

\multicolumn{1}{c|}{\multirow{-11}{*}{INbreast}} & \textbf{ Proposed cGAN}             & \textbf{68.69}                                      & \textbf{94.07}                                       & \textbf{92.11}                                       & \textbf{52.31}                                     & \textbf{87.03}                                     & \textbf{84.55}\\ 

\multicolumn{1}{c|}{}                           & Dhungel et al.\cite{dhungel2015deep}                          & $-$                                               & $-$                                                & $90.00$                                       & $-$                                          & $-$                                              & $-$         \\

\multicolumn{1}{c|}{}                           & Cardoso et al.\cite{cardoso2017mass}                          & $-$                                               & $-$                                                & $90.00$                                        & $-$                                          & $-$                                              & $-$         \\

\multicolumn{1}{c|}{}                           & Zhu et al.\cite{zhu2018adversarial}& $-$                                               & $-$                                                & $90.97$                                       & $-$                                          & $-$                                              & $-$         \\

\multicolumn{1}{c|}{}                           &  Al-antari et al.\cite{al2018fully}                          & $-$                                               & $-$                                                &  $92.69$                                       & $-$                                          & $-$                                              & $86.37$         \\\hline

\end{tabular}
}
\end{table}

The results depicted in Table~\ref{tabDiceIoU} are divided in two sections, one for our private dataset and another for the INbreast dataset. Note that all models are trained on the private dataset, and then tested using our private dataset as well as the INbreast dataset without fine tuning. 

According to the results, our method outperforms the compared state-of-the-art methods in all cases except for the IoU computed on tight crops of our private dataset. The SLSDeep approach yielded the best IoU ($79.93\%$), whereas our method yielded the second best result ($79.87\%$).

All models yielded their worst segmentation results for full mammograms compared to other frame inputs, which is logical taking into account the difficulties stated earlier in this section. Most of the models have obtained their best results for the tight frame crops except for CRFCNN and our proposal, which yielded their best results for loose frame crops. However, the good results for tight crops may be due to the imbalance of tumor/non-tumor pixels, since the former class is present in more than $90\%$ of the image area. The learning can be biased towards this class, which makes rough solutions (almost everything is tumor) to provide very high ranks of performance. Loose frame crops, on the contrary, have a more balanced proportion of pixels for both classes, which makes them ideal to learn and evaluate the model on a realistic situation: it is more convenient for radiologists to provide a fast frame drawing around the breast tumor rather than a tight frame.


Comparing the general results for both datasets, most methods performed better on INbreast rather than on private dataset with loose and tight framing. This effect can be explained by the fact that INbreast provides more detailed ground truths, which leads to better testing results, despite all network training has been conducted on our private dataset.

In general, our proposal has performed well in terms of both Dice and IoU metrics. For private dataset, in Dice/Loose frame column, our model's percentage ($89.99\%$) is almost $9\%$ above the second best model, SegNet-VGG16 ($81.09\%$). In the IoU/Loose frame column, our model's percentage ($81.81\%$) is almost $10\%$ above the second best model, cGAN-ResNet101 ($71.91\%$). For INbreast dataset, our Loose frame results for Dice and IoU are again the best ($94.07\%$, $87.03\%$), where cGAN-ResNet101 is the second best model for both metrics ($87.19\%$, $77.51\%$). Thus, our model provides an improving of $7\%$ and $10\%$, respectively. The fact that the second best results are obtained by the cGAN-ResNet101 model indicates that the adversarial network really helps in training the generative network. In turn, the results obtained by the cGAN-ResNet101 (Dice Loss) mixture model are in-between the cGAN-ResNet101 and our proposal, since the Dice loss term substitution improves the accuracy of tumor segmentations. 

For the INbreast dataset, we have included the results mentioned in four related papers~\cite{dhungel2015deep}, \cite{cardoso2017mass}, \cite{zhu2018adversarial} and \cite{al2018fully}. For these methods, we could not compute the metrics for all columns, since they have not released their source code. Our method outperformed the first three papers under similar framework conditions. However, \cite{al2018fully} yielded better results for dice ($92.69\%$) and IoU ($86.37\%$) than our model in the Tight frame columns. Our results in the Loose frame columns surpass their results. For a fair comparison, however, it should be checked how the referenced methods would perform on loose frame crops.

\begin{figure}[h!]
\centering
\includegraphics[width=0.75\textwidth, height=0.30\textheight]{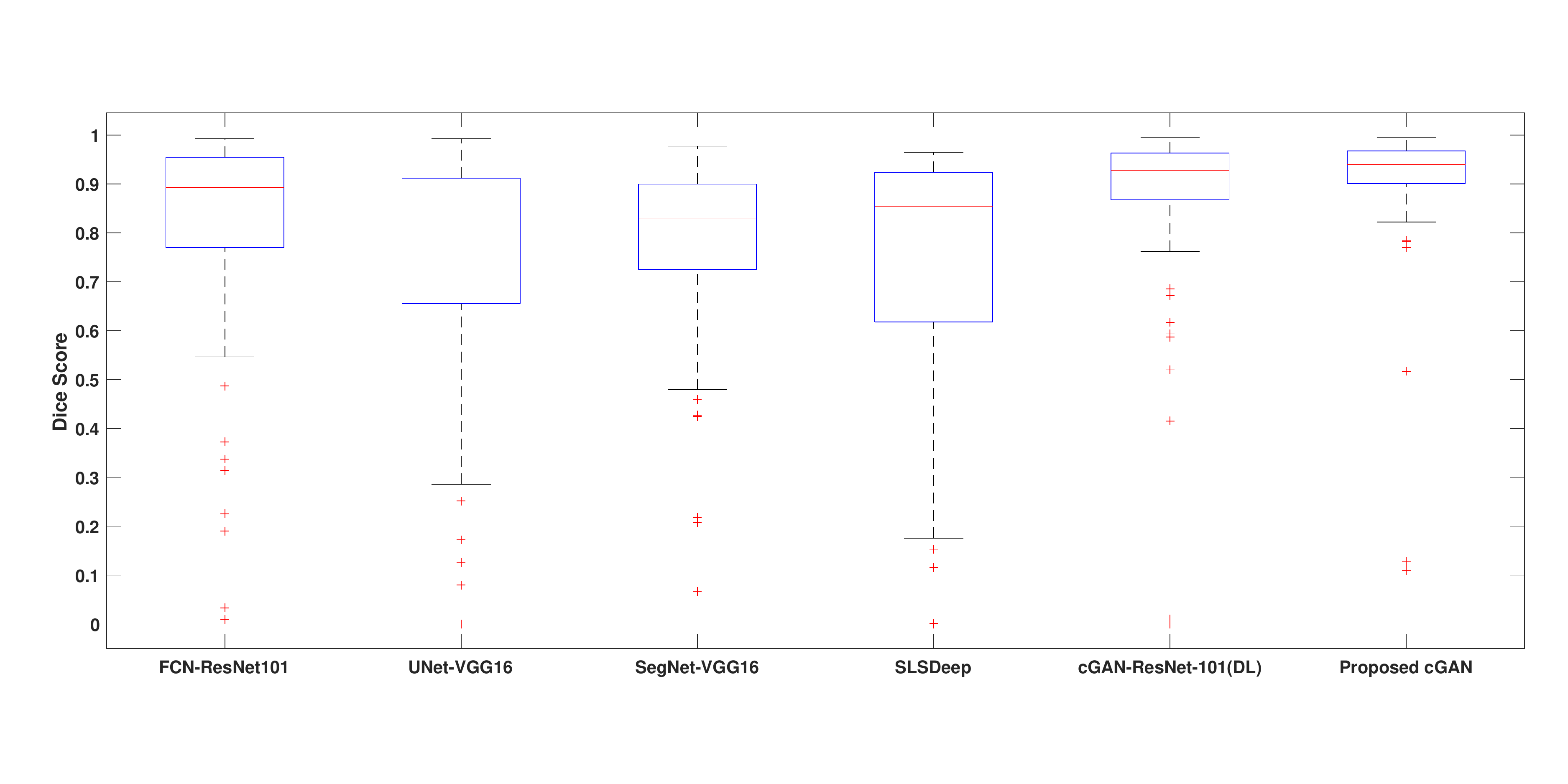}\\
\includegraphics[width=.75\textwidth, height=0.30\textheight]{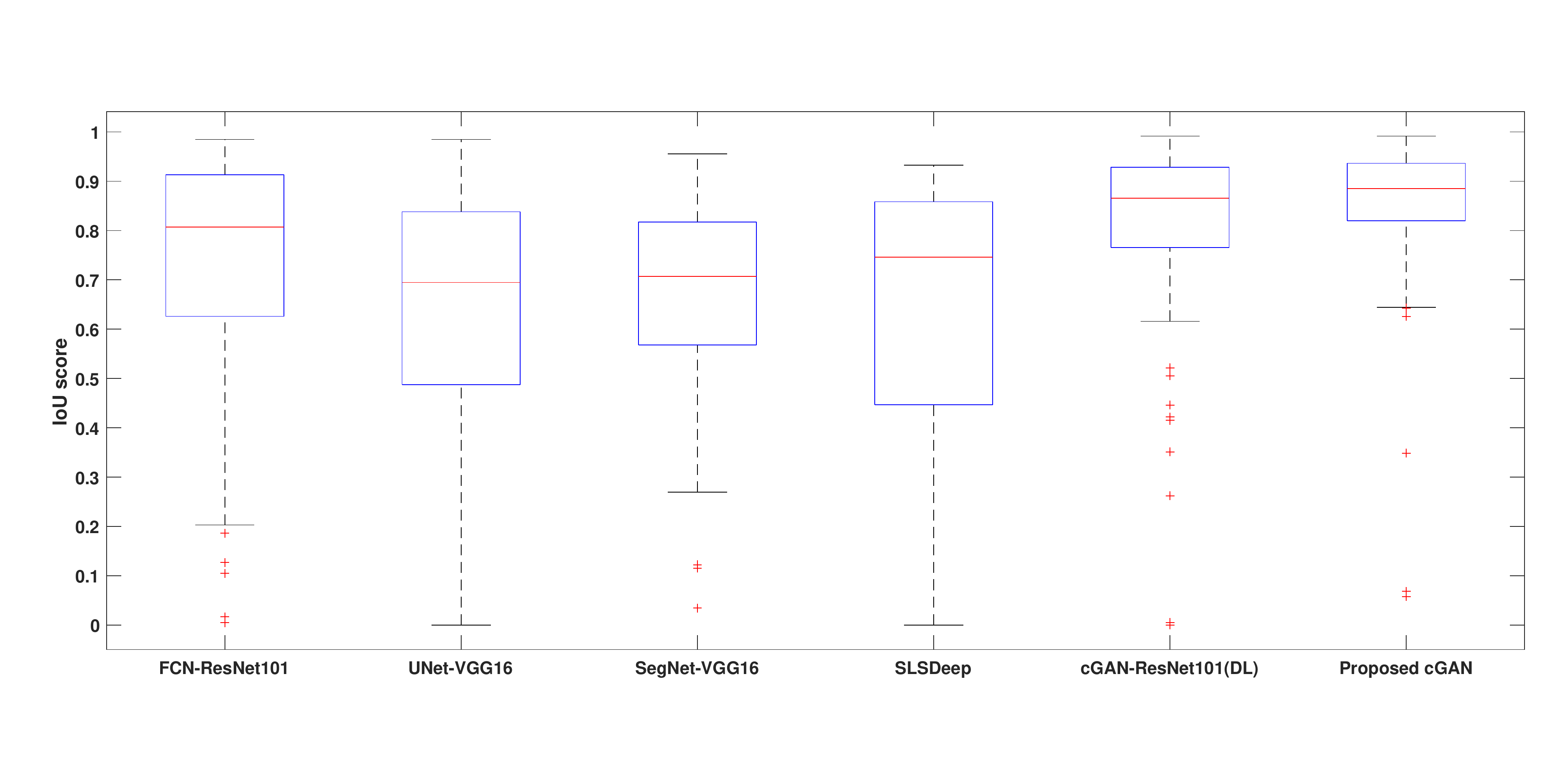}
\caption{Boxplot of dice (Top) and IoU(Bottom) score over five models  compared to proposed cGAN on loose frames of the test subset of INbreast dataset (106 samples).}
\label{fig:lossfunction}
\end{figure}

The box-plot in Figure \ref{fig:lossfunction} shows Dice and IoU values obtained for the 106 testing samples from INbreast dataset with loose frames using FCN-ResNet101, Unet-VGG16, SegNet-VGG16, SLSDeep, cGAN-ResNet101 and proposed cGAN. The two models based on cGAN provide small ranges of Dice and IoU values. For instance, the proposed cGAN is in the range 0.89 to 0.93 for Dice coefficient and 0.80 to 0.91 for IoU values, while other deep segmentation methods, SLSDeep, Unet-VGG16 and FCN-ResNet101, show a wider range of values. Moreover, there are many outliers in the results for the segmentation based on the cGAN using pre-trained ResNet101 layers, while using our cGAN trained from scratch there are few number of outliers.

The high Dice and IoU metrics obtained by our model empirically support our hypothesis that it achieves accurate tumor segmentation. In Fig.~\ref{fig:segment2x3examples}, we show some examples of our model's segmentations using two tumors from the INbreast dataset by applying all three cropping strategies. For each experiment, we show the original ROI image and the comparison of predicted and ground truth mask, color coded to mark up the true positives (TP:yellow), false negatives (FN:red), false positives (FP:green) and true negatives (TN:black). For the full mammogram, the ROI image (1) is an example of good segmentation, since yellow and black pixels depict a high degree of confidence between predicted and real masks. On the contrary, the ROI image (2) is an example of poor segmentation, since red pixels mark up a high portion of the breast tumor area that has been misclassified as healthy area (FN). At the same time, a tiny region of green pixels shows the misclassification of healthy tissue as breast tumor area (FP). Nevertheless, even in this second segmentation, there is a very high rate of black pixels (TN), which indicates that the model easily recognizes non-tumor areas.

In the loose frame segmentations (middle row), specially with example (2), the results contain very few FN and FP pixels. For example (1), a modest amount of green pixels indicate that our model expands the tumor segmentation beyond its respective ground truth. In the tight frame crops (bottom row), besides the green areas, our model also has missed some tumor areas \emph{i.e.}, the red pixels (FN). The mistaken areas (red and green) are mostly around the tumor borders, since these areas have a mixture of healthy and unhealthy cells. At the same time, the inner part of the tumor as well as the image regions outside of tumors are properly classified, which indicates the stability of our model.

\begin{figure*}[htp]
\centering
\includegraphics[width=1.0\textwidth, height=0.75\textheight]{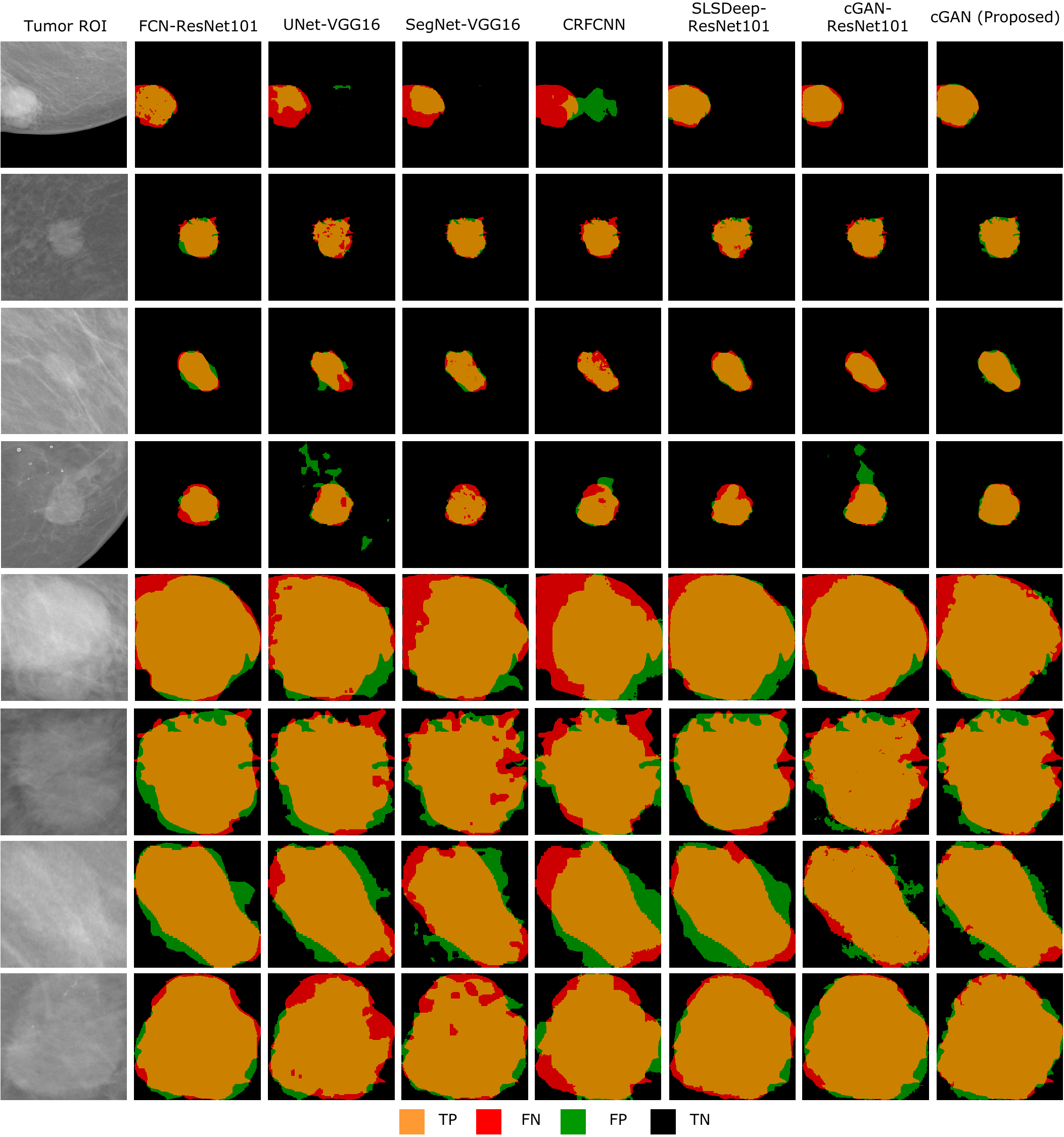}
\caption{Segmentation results of seven models with the INbreast dataset and two cropping strategies: loose frame (the first four rows) and tight frame (the last four rows). (Col 1) original images, (Col 2) FCN-ResNet101, (Col 3) UNet-VGG16, (Col 4) SegNet-VGG16, (Col 5) CRFCNN, (Col 6) SLSDeep, (Col 7) cGAN-ResNet101, and (Col 8) proposed cGAN.}
\label{fig:segment8x8examples}
\end{figure*}

Fig.~\ref{fig:segment8x8examples} shows a comparison between our and other six segmentation models, which worked on loose and tight frame crops using four tumors from the INbreast dataset. For the loose frame cases (four top rows), our method clearly outperforms the rest for all tumors except for the second one, where the majority of models provided a similar degree of accuracy. In these four tumors, UNet-VGG16 and CRFCNN provided the worst results. Moreover, cGAN-ResNet101 also performed bad in the fourth example.

For the tight frame cases (four bottom rows), our method also provides the lowest degrees of FN and FP compared to the rest of the models. Our cGAN and the cGAN-ResNet101 model yield irregular borders compared to FCN-ResNet101 and SLSDeep, since GAN models strive for higher accuracy on edges. However, in the third tight frame sample (seventh row), both cGAN-ResNet101 and our proposal generated an irregular border that slightly differs from the smooth ground truth border, which results in lower segmentation accuracy around the edges. Although the rest of the models generate smoother borders, the resulting segmentations may differ from the ground truth significantly.

From the experimental results, it can be concluded that the proposed breast tumor segmentation method is the most effective to date compared to the currently available state-of-the-art methods. However, our method needs a loose crop around the tumor to obtain a proper segmentation, which can be done by the SSD model. Our segmentation model contains about $13,607,043$ parameters for tuning the generator part in the cGAN network. In addition, our method is fast in both training \emph{i.e.}, around $30$ seconds per epoch ($220$ loose frames) and predicting, around $7$ images per second. That is $7$ to $8$ times faster than the segmentation method proposed in \cite{al2018fully} and 10 to 15 times faster than the FCN model.

\subsection{Shape Classification Experiments}

For validating the tumor shape classification performance, we computed the confusion matrix and the overall classification accuracy on the test set of the DDSM dataset. This set contains 292 images divided into 126, 117, 31 and 18 for irregular, lobular, oval and round classes, respectively.

For a quantitative comparison, we compared our model with three state-of-the-art tumor shape classification methods~\cite{singh2018conditional,kisilev2015semantic,kim2018icadx}. The three methods were evaluated on the DDSM dataset. 

\begin{table}[htp]
\centering
    
\caption{Confusion matrix of the tumor shape classification of testing samples of the DDSM dataset.}
\label{Table2}
\begin{tabular}{|p{2.2cm}|p{1.4cm}|p{1.4cm}|p{1.4cm}|p{1.4cm}|p{0.9cm}|}
		 \hline
\begin{tabular}[c]{@{}c@{}} Prediction / \\Ground Truth \end{tabular} & {\cellcolor[HTML]{E5E7E9}Irregular} & {\cellcolor[HTML]{E5E7E9}Lobular} & {\cellcolor[HTML]{E5E7E9}Oval} & {\cellcolor[HTML]{E5E7E9}Round} &  {\cellcolor[HTML]{E5E7E9}Total} \\  \hline

{\cellcolor[HTML]{E5E7E9  }Irregular}   & {\cellcolor[HTML]{ 08eeeb }$96~(76\%)$}        &{\cellcolor[HTML]{FE030F}$30$}      &$0$   &$0$     &$126$  \\ \hline

{\cellcolor[HTML]{E5E7E9}Lobular}  & {\cellcolor[HTML]{FE030F}$33$}  & {\cellcolor[HTML]{08EEE7}$83~(71\%)$} & {\cellcolor[HTML]{F66107}$1$}    & $0$     & $117$    \\ \hline

{\cellcolor[HTML]{E5E7E9}Oval}   & $0$    & {\cellcolor[HTML]{F66107}$1$}  & {\cellcolor[HTML]{08C8EE}$26~(84\%)$}   & {\cellcolor[HTML]{FE030F}$4$}    &  $31$   \\ \hline

{\cellcolor[HTML]{E5E7E9}Round}    & $0$    &  {\cellcolor[HTML]{F66107}$1$}     & {\cellcolor[HTML]{F66107}$1$}    & {\cellcolor[HTML]{ 08B6EE}$16~(89\%)$}    &  $18$    \\ \hline 

\end{tabular}
\end{table}

However, The DDSM dataset does not have the ground truth binary masks for the breast tumor segmentation.
Thus, we applied active contours~\cite{Akram2015}, which was also used in our previous work~\cite{singh2018conditional}, to generate the ground truths of the breast tumor regions that were cropped by radiologists. Previously,~\cite{kisilev2015semantic} also used active contours~\cite{lankton2008localizing} to generate the ground truths in a similar fashion. In addition, for reliable performance results, we used a stratified 5 fold cross validation with $50$ epochs per fold.

In Table~\ref{Table2}, the proposed method yielded around $73\%$ of classification accuracy for irregular and lobular classes. This result is logical, since both lobular and irregular shapes have similar irregular boundaries. In turn, our model yielded classification accuracies of $84\%$ and $89\%$ for oval and round shape classes, respectively.

\begin{figure}[htp]
\centering
\includegraphics[width=0.7\columnwidth]{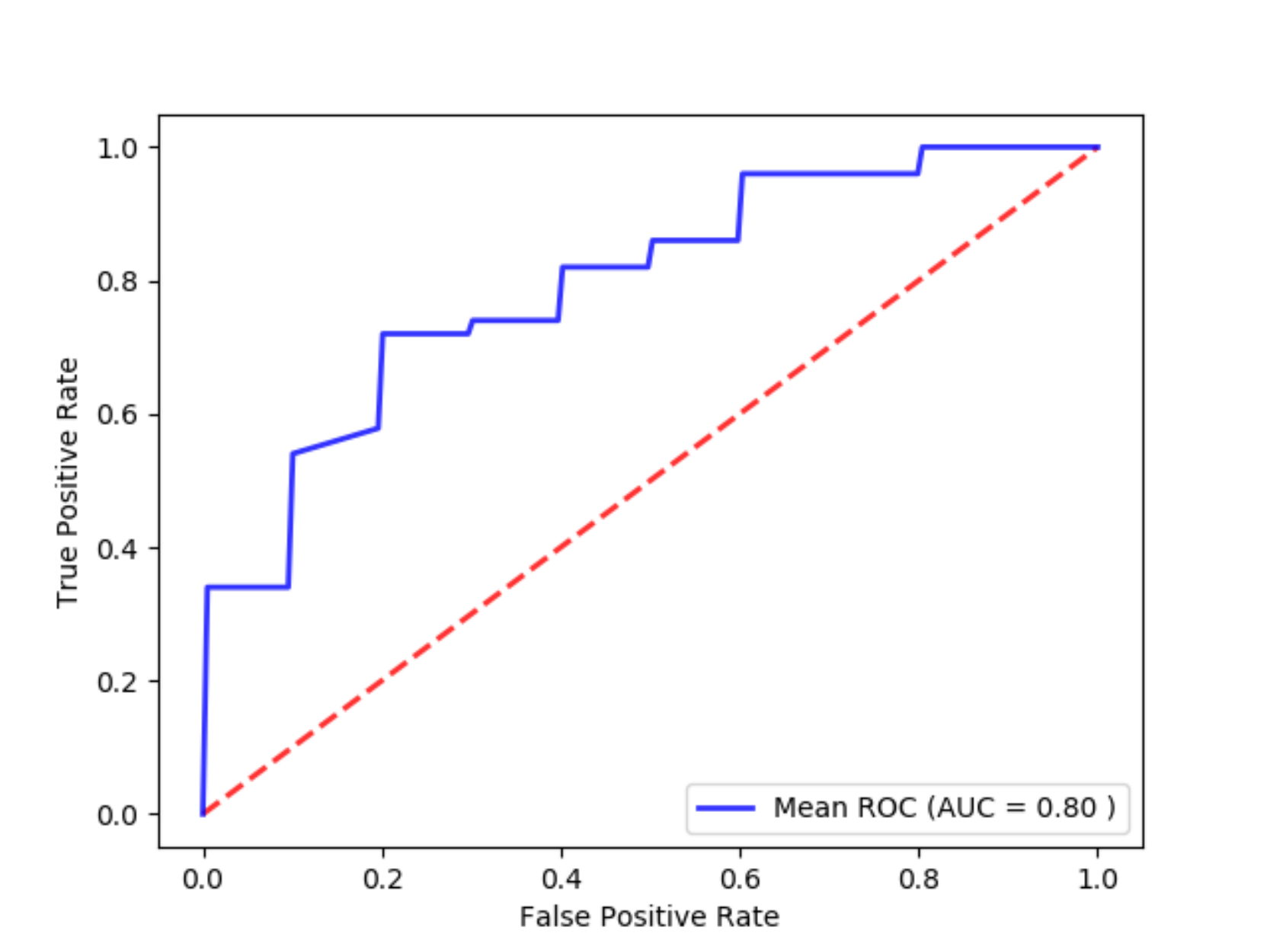}
\caption{Mean ROC curve of 5 folds, for TPR and FPR from shape classification result of 292 test images from DDSM dataset.}
\label{fig:Roc}
\end{figure}

We have computed the overall accuracy of each method by averaging the correct predictions (i.e., true positive) of the four classes, weighted with respect to the number of samples per class. As shown in Table~\ref{table3}, our classifier yields an overall accuracy of $80\%$, outperforming the second best results~\cite{kim2018icadx,singh2018conditional} by $8\%$. In turn, Multi-task CNN~\cite{kim2018icadx} based on a pre-trained VGG-16 yielded the worst overall accuracy ($66\%$), probably because the input mammograms are gray-scale images, while the VGG-16 network was trained on color-scale images. In addition, Fig.~\ref{fig:Roc} shows ROC curve illustrating that our model attained AUC about 0.8.



Furthermore, the proposed shape descriptor contains 767,684 parameters, which can be trained in less than a second per epoch, and predict in about 6 milliseconds per image. 


\begin{table}[htp]
\centering
\caption{Shape classification overall accuracy with the DDSM dataset resulting from ~\cite{kisilev2015semantic,kim2018icadx, singh2018conditional} and our model. Best result is marked in bold.}
\label{table3}
\scalebox{0.9}{
\begin{tabular}{|c|c|}
\hline
\cellcolor{gray!25} Methods & \cellcolor{gray!25}Accuracy (\%) \\ \hline 
       Kisilev et al. (SSVM) \cite{kisilev2015semantic} &   ~$71 $         \\ 
      Kim et al. (Multi-task CNN) \cite{kim2018icadx} &    ~$66$          \\
        Kim et al. (ICADx) \cite{kim2018icadx}&  ~$72$          \\ 
        Singh et al. \cite{singh2018conditional} &   ~$72$         \\ 
       \textbf{Proposed} &  \textbf{~80}          \\ \hline
\end{tabular}
}
\end{table}

\subsection{Shape Features Correlation to Breast Cancer Molecular Subtypes}

\begin{table}[htp]
	\centering
	\caption{Distribution of breast cancer molecular subtypes samples from the hospital dataset with respect to its predicted mask shape.}
	\label{Table4}
    \scalebox{1.0}{
 	\begin{tabular}{|p{2.9cm}|p{1.4cm}|p{1.4cm}|p{1.4cm}|p{1.4cm}|p{0.8cm}|}
		\hline
		\begin{tabular}[c]{@{}c@{}} Shape Classes / \\Molecular Subtypes \end{tabular} & {\cellcolor[HTML]{E5E7E9}Irregular} & {\cellcolor[HTML]{E5E7E9}Lobular} & {\cellcolor[HTML]{E5E7E9}Oval} & {\cellcolor[HTML]{E5E7E9}Round} &  {\cellcolor[HTML]{E5E7E9}Total} \\ \hline 
		{\cellcolor[HTML]{E5E7E9  }Luminal-A}                                                                    & {\cellcolor[HTML]{ F39C12 }$~~~~67~$}        &{\cellcolor[HTML]{ F39C12 }$~~~~29$}      & $~~~~10$   & $~~~~17$     &  $123$    \\ \hline
		{\cellcolor[HTML]{E5E7E9}Luminal-B }                                                                    & {\cellcolor[HTML]{F39C12 }$~~~~58$}        & {\cellcolor[HTML]{F39C12  }$~~~~24$}	      & $~~~~14$    & $~~~~11$     & $107$    \\ \hline
		{\cellcolor[HTML]{E5E7E9}Her-2}                                                                       & $~~~~6$         & $~~~~4$       &  {\cellcolor[HTML]{43C349}$~~~~8$}   & {\cellcolor[HTML]{43C349  }$~~~~15$}    &  $33$   \\ \hline
		{\cellcolor[HTML]{E5E7E9}Basal-like}                                                                   & $~~~~5$         & $~~~~10$      & {\cellcolor[HTML]{43C349  }$~~~~9$}    & {\cellcolor[HTML]{43C349}$~~~~13$}    &  $37$    \\ \hline 
	\end{tabular}}
\end{table}

Tumor shape could play an important role to predict the breast cancer molecular subtypes \cite{tamaki2011correlation}. Thus, we have computed the correlation between breast cancer molecular subtypes classes of our in-house private dataset with the four shape classes. As shown in Table~\ref{Table4}, most of Luminal-A and -B samples (i.e., 96/123 and 82/107 for Luminal-A and -B, respectively) are mostly assigned to irregular and lobular shape classes. In turn, oval and round tumors give indications to the Her-2 and Basal-like samples, (i.e., 23/33 and 22/37 for Her-2 and Basal-like, respectively). Moreover, some images related to Basal-like are moderately assigned to the lobular class. Afterwards, from the visual inspection, if the tumor shape is irregular or lobular then radiologist can suspect that it belongs to the Luminal group. In turn, if the tumor shape is round or oval then it is more probable that the tumor is a Her-2 or Basal-like \cite{tamaki2011correlation}. Therefore, this study shows the importance of tumor shape, which can be considered as a key feature to distinguish between different malignancies of breast cancer.

\subsection{Limitations}

\begin{figure}[htp]
\centering
\includegraphics[width=0.6\columnwidth]{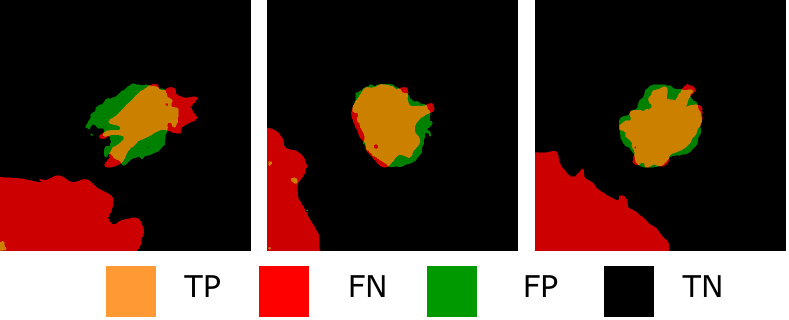}
\caption{Three mis-segmented of non-full tumor shapes with INbreast dataset. The red part in the down-left border.}
\label{fig:limitation}
\end{figure}

For the segmentation stage, our model has two limitations. The first one is that a prior information about the tumor location must be provided in order to center the tumor inside a loose frame crop for obtaining the best accuracy. To alleviate this requisite, we propose to use the deep SSD model to localize the tumor region to have a complete automatic process, instead of a radiologist manually detecting the tumor region. The second limitation is that our cGAN model is prepared to segment tumors fully contained in the ROI, otherwise, the model fails to segment it. As shown in Fig.~\ref{fig:limitation}, we found three samples that are mis-segmented because they contained two tumors, the one in the center, which is properly segmented, and another that is shown partially in the left-down border of the image, which is wrongly ignored as non-tumor region (FN). When the bigger tumor is located in the center of the crop, nevertheless, it is correctly segmented.

To classify the tumor shape, we depend only on the DDSM dataset to train our model, since it is the only public dataset that has the shape classification information. Thus, more databases containing more samples are required to improve the classification accuracy of four shape classes.

To study the molecular subtypes of breast cancer, Her-2 and Basal-like classes have less samples compared to the other two classes, Luminal-A and Luminal-B. Indeed, we used a weighted loss function to train our shape classification model in order to make a balance between the four classes. However, we anticipate that, by increasing the samples related to the Her-2 and Basal-like classes, we will improve the prediction of molecular subtypes from tumor shape information.

\section{Conclusion}
\label{sec:conclusion}
In this paper, we propose a two stage breast tumor segmentation and classification method, which first segments the breast tumor ROI using a cGAN and then classify its binary mask using a CNN based shape descriptor.

The segmentation results reveal the importance of the adversarial network in the optimization of the generative network. cGAN-ResNet101 shows an improvement of about $1\%$ to $3\%$ in both Dice and IoU metrics in comparison to the other non-GAN methods. In turn, the proposed method yields an increment of about $10\%$ over the results of cGAN-ResNet101 by training our model from scratch, and replacing the $L1$-norm with the dice loss using loose frame crop on the given datasets. The breast tumor segmentation from full-mammograms yields low segmentation accuracy for all models including the proposed cGAN. For the tight frame crop, the proposed cGAN yields similar or better segmentation accuracy compared to the other methods.

The classification results show that our second stage properly infers the tumor shape from the binary mask of the breast tumor, which was obtained from the first stage (cGAN segmentation). Hence, we have empirically shown that our CNN is focusing its learning on the morphological structure of the breast tumor, while the rest of approaches (\cite{kisilev2015semantic}, \cite{  kim2018icadx}, \cite {kisilev2016medical}, \cite{ren2015faster}) rely on the original pixel variations of the input mammogram to make the same inference. Moreover, in \cite{al2018fully} they used a hybrid strategy in which they include the pixel variability within the mask of breast tumor region to retain the intensity and texture information. However, the superior performance obtained by our method supports our initial idea that the second stage CNN can reliably recognize the tumor shape based only on morphological information.

Furthermore, this paper provided a study of correlation between the tumor shape and the molecular subtypes of the breast cancer. Most samples of the Luminal-A and -B group are assigned to irregular shapes. In turn, the majority of Her-2 and Basal-like samples are assigned to regular shapes (e.g., oval and round shapes). That gives an indication that the tumor shape can be considered for inferring the molecular subtype of the tumor.

Future work aims at refining our multi-stage framework to detect other breast tumor features (\emph{i.e.}, margin type, micro-calcifications), which will be integrated into a more comprehensive diagnostic to compute the degree of malignancy of the breast tumors.





\section*{Conflict of interest} 
The authors declare that there is no conflict of interest.

\bibliography{mybibfile}

\end{document}